\documentclass[10pt,journal,compsoc]{IEEEtran}

\usepackage[pdftex]{graphicx}    
\usepackage{cite}
\usepackage{booktabs}
\usepackage{amsmath}
\usepackage{algorithm}
\usepackage[noend]{algpseudocode}
\usepackage[
singlelinecheck=false 
]{caption}

\makeatletter
\def\BState{\State\hskip-\ALG@thistlm}
\makeatother

\usepackage{fixltx2e}
\usepackage{xcolor}
\def\SPSB#1#2{\rlap{\textsuperscript{\textcolor{red}{#1}}}\SB{#2}}

\def\SB#1{\textsubscript{\textcolor{blue}{#1}}}

\newcommand{\subparagraph}{}
\usepackage{titlesec}

\titleformat{\paragraph}
{\normalfont\normalsize\bfseries}{\theparagraph}{1em}{}
\titlespacing*{\paragraph}
{0pt}{3.25ex plus 1ex minus .2ex}{1.5ex plus .2ex}

\begin{document}

\title{Robot localization in a mapped environment using Adaptive Monte Carlo algorithm}

\author{Sagarnil Das}

{}
\IEEEtitleabstractindextext{%

\begin{abstract}
Localization is the challenge of determining the robot's pose in a mapped environment. This is done by implementing a probabilistic algorithm to filter noisy sensor measurements and track the robot's position and orientation. This paper focuses on localizing a robot in a known mapped environment using Adaptive Monte Carlo Localization or Particle Filters method and send it to a goal state. ROS, Gazebo and RViz were used as the tools of the trade to simulate the environment and programming two robots for performing localization.
\end{abstract}

}

\maketitle
\IEEEdisplaynontitleabstractindextext
\IEEEpeerreviewmaketitle
\section{Introduction}
\label{sec:introduction}

\IEEEPARstart{T}{he} localization problem is of utmost importance in the real world as this gives us a probabilistic estimate of the robot's current position and orientation. So, it is very obvious that without this knowledge, the robot won't be able to take effective decisions and take sound actions if it doesn't know where it is located in the world. There are 3 different types of localization problems.

a) \textbf{Local Localization:} This is the easiest localization problem. It is also known as position tracking. In this problem, the robot knows its initial pose and the localization challenge entails estimating the robot's pose as it moves out in the environment. This problem is not trivial as there is always some uncertainty in robot motion. However, the uncertainty is limited to regions surrounding the robot.

b) \textbf{Global Localization:} This is a more complicated localization problem. In this case, the robot's initial pose is unknown and the robot must determine its pose relative to the ground truth map. The amount of uncertainty is much higher.

c) \textbf{The kidnapped robot problem}: This is the most challenging localization problem. This is just like the global localization problem, except that the robot may be kidnapped at any time and moved to a new location on the map.

\begin{figure}[thpb]
      \centering
      \includegraphics[width=\linewidth]{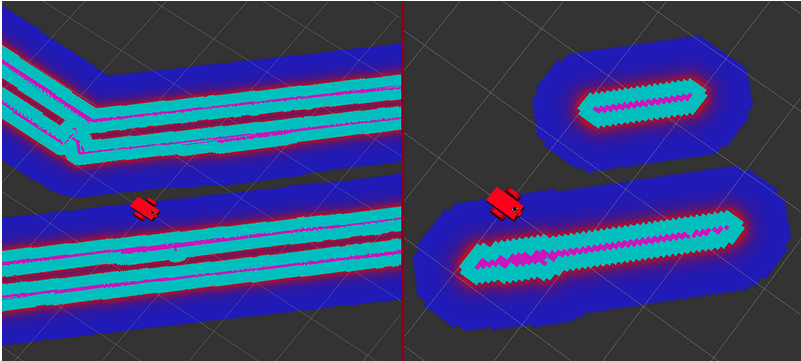}
      \caption{Robot Localization}
      \label{fig:robot_localization_costmap}
\end{figure}

Two robots were developed and tested in a simulation environment. Both the robots successfully localized themselves and navigated the maze using Adaptive Monte Carlo algorithm (AMCL). The benchmark robot's URDF was given as a part of the project whereas the second robot was build independently. The benchmark robot was called \textit{UdacityBot} and the 2nd robot was called \textit{SagarBot}. The world-map is called 'Jackal-Race' and was created by Clearpath Robotics.\cite{jackalrace}

\section{Background}
For the localization problem, a wide range of algorithms are available ranging from Monte Carlo Localization, Extended Kalman Filter to Markov and finally Grid Localization. The Monte Carlo Localization algorithm or MCL, is the most popular localization algorithms in robotics. After MCL is deployed, the robot will be navigating inside its known map and collect sensory information using RGB camera and range-finder sensors. MCL will use these sensor measurements to keep track of the robot's pose. MCL is often referred to as Particle Filter Localization, since it uses particles to localize the robot. These particles are virtual elements that resembles the robot. Each particle has a position and orientation and it represent a guess where the robot might be located. These particles are re-sampled each time the robot moves and senses its environment. For this project, a modified version of this algorithm known as \textbf{'Adaptive Monte Carlo Localization'} was used because this modified algorithm dynamically adjusts the number of particles over a period of time, as the robot navigates around the map, hence making the process more cost effective.

Some of the current challenges of localization involves:

a) It doesn't operate well in an unmapped environment.

b) In the multi robot localization problem, which involves a team of robots which simultaneously seek to determine their poses in a known environment, dependencies are created in the pose estimates of individual robots that pose major challenges for the design of the estimator.\cite{particle_filters}

\subsection{Kalman Filters}

The Kalman filter is an estimation algorithm that is very prominent in controls. It is used to estimate the value of a variable in real time as the data is being collected. This variable can represent the position or velocity of the robot, or even the temperature of a process. 

The reason that the Kalman filter is so net worthy is because it can take data with a lot of uncertainty or noise in the measurement and provide a very accurate estimate of the real values in a very short time. Unlike other estimation algorithms, it doesn't depend on a lot of data to come in order to calculate an accurate estimate. The Kalman filter was invented during the Apollo program. It was used to help Apollo enter the orbit of the moon. Since its success with the Apollo program, Kalman filter has become one of the most practical algorithms in the field of control engineering. 

The Kalman filter works cyclically between two steps. The Kalman filter produces an estimate of the state of the system as an average of the system's predicted state and of the new measurement using a weighted average. The purpose of the weights is that values with better (i.e., smaller) estimated uncertainty are "trusted" more. The weights are calculated from the covariance, a measure of the estimated uncertainty of the prediction of the system's state. The result of the weighted average is a new state estimate that lies between the predicted and measured state, and has a better estimated uncertainty than either alone. This process is repeated at every time step, with the new estimate and its covariance informing the prediction used in the following iteration. This means that the Kalman filter works recursively and requires only the last "best guess", rather than the entire history, of a system's state to calculate a new state. 

The linear Kalman filter assumes that the output is proportional to the input and hence it can be only applied to linear systems. This limitation is overcome with Extended Kalman filters, which can be applied to non-linear systems, which is more applicable in robotics as real world systems are more often non-linear than linear. Also Linear Kalman filter assumes that both the prior and the posterior follows a unimodal Gaussian distribution. But in real world, that is seldom the case. Extended Kalman filters overcome this problem by linear approximation of the posterior distribution after a non-linear transformation.\cite{kalman_filters}

\subsection{Particle Filters}
The Monte Carlo Localization or Particle Filters uses virtual particles to estimate a robot's pose. With MCL, particles are initially spread uniformly and randomly throughout the entire map. Just like the robot, each particle has a x-y coordinate and an orientation vector. So each of these particles represent the hypothesis of where the robot might be. In addition to the 3d vector, particles are assigned a weight. The weight of a particle is the difference between the robot's actual pose and the particle's predicted pose. The importance of a particle is dependent on its weight. The bigger the particle, more accurate it is.

Particles with larger weights are more likely to survive during the re-sampling process. After the re-sampling process, particles with significant weights are more likely to survive whereas others are more likely to die. Finally after several iterations of the algorithm and after different stages of re-sampling, particles will converge and estimate the robot's pose. 

The MCL algorithm estimates the posterior distribution of a robot's position and orientation based on sensory information. This process is known as \textit{Bayes Filter}. Using a Bayes filtering approach, the state of a dynamical system can be estimated from the sensor measurements.

The MCL algorithm is composed of two main sections represented by two for-loops.

\begin{algorithm}
\caption{MCL algorithm}\label{euclid}
\begin{algorithmic}[1]
\Procedure{MCL}{$x_{t-1}, u_{t}, z_{t}$}
\State $\textit{$X_{t}$} \gets \textit{$\phi$}$
\BState \emph{for m=1 to M loop}:
\State $x\SPSB{[m]}{t}  \gets \textit{$MotionUpdate(u_{t}, x\SPSB{[m]}{t-1} $)}$
\State $w\SPSB{[m]}{t}  \gets \textit{$SensorUpdate(z_{t}, x\SPSB{[m]}{t} $)}$
\State $X_{t} \gets \textit{$X_{t} + <x\SPSB{[m]}{t} + w\SPSB{[m]}{t}> $)}$
\BState \emph{end for}

\BState \emph{for m=1 to M loop}:
\State $\textit{draw x\SPSB{[m]}{t} with probability $\propto$ w\SPSB{[m]}{t}} $

\State $X_{t} \gets \textit{$X_{t} + x\SPSB{[m]}{t} $)}$
\BState \emph{end for}
\BState \emph{return $X_{t}$}
\EndProcedure
\end{algorithmic}
\end{algorithm}

The first section is the motion and sensor update and the second one is the re-sampling process. Given a map, MCL is to determine the robot's pose represented by the belief ($X_{t}$). 

At each iteration, the algorithm takes the previous belief ($X_{t-1}$), the actuation command ($u_{t}$) and the sensor measurement ($z_{t}$) as input. Initially, the belief is obtained by randomly generating m particles. Then in the first loop, the hypothetical state is computed whenever the robot moves. Following, the particles' weight is computed using the latest sensor measurement. Now motion and measurements are both added to the previous state.

In the second section of the MCL, a simple sampling process happens. Here, the particles with high probability survive and are re-drawn in the next iteration, while the others die.

Finally, the algorithm outputs the new belief and another cycle of iteration starts implementing the next motion by reading the new sensor measurements.

\subsection{Comparison / Contrast}
There are certain significant benefits of Monte Carlo Localization over Extended Kalman Filter algorithm. Firstly, MCL is easy to code. Secondly, MCL represents non-Gaussian distribution and can approximate any other practical important distribution. This means MCL is unrestricted by a linear Gaussian state based assumption as in the case of EKF. This allows MCL to model a much greater variety of environments specially since the real world cannot be always modeled by Gaussian distributions. Thirdly, in MCL, the computational memory and the resolution of the solution can be controlled by changing the number of particles distributed uniformly and randomly throughout the map.

The general difference between MCL and EKF is described in the table below.

\begin{figure}[thpb]
      \centering
      \includegraphics[width=\linewidth]{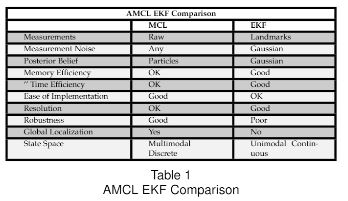}
      \caption{EKF vs MCL comparison}
      \label{fig:ekf_vs_mcl}
\end{figure}

So MCL is generally more advantageous than EKF and hence MCL will be implemented for the purpose of this project.

\section{Simulations}
The simulations were done in a ROS environment using Gazebo and RViz. The navigation stack\cite{navigation_stack} can be visualized as follows:

\begin{figure}[thpb]
      \centering
      \includegraphics[width=\linewidth]{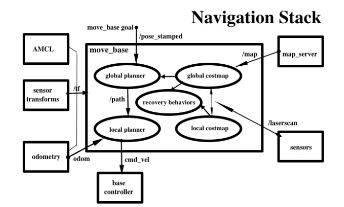}
      \caption{Navigation Stack}
      \label{fig:navigation_stack}
\end{figure}

The whole simulation process was carried out on both UdacityBot and SagarBot. The algorithm and map visualization process were done in RViz. We can also see the robot in action inside the Gazebo simulation environment.

\begin{figure}[thpb]
      \centering
      \includegraphics[width=\linewidth]{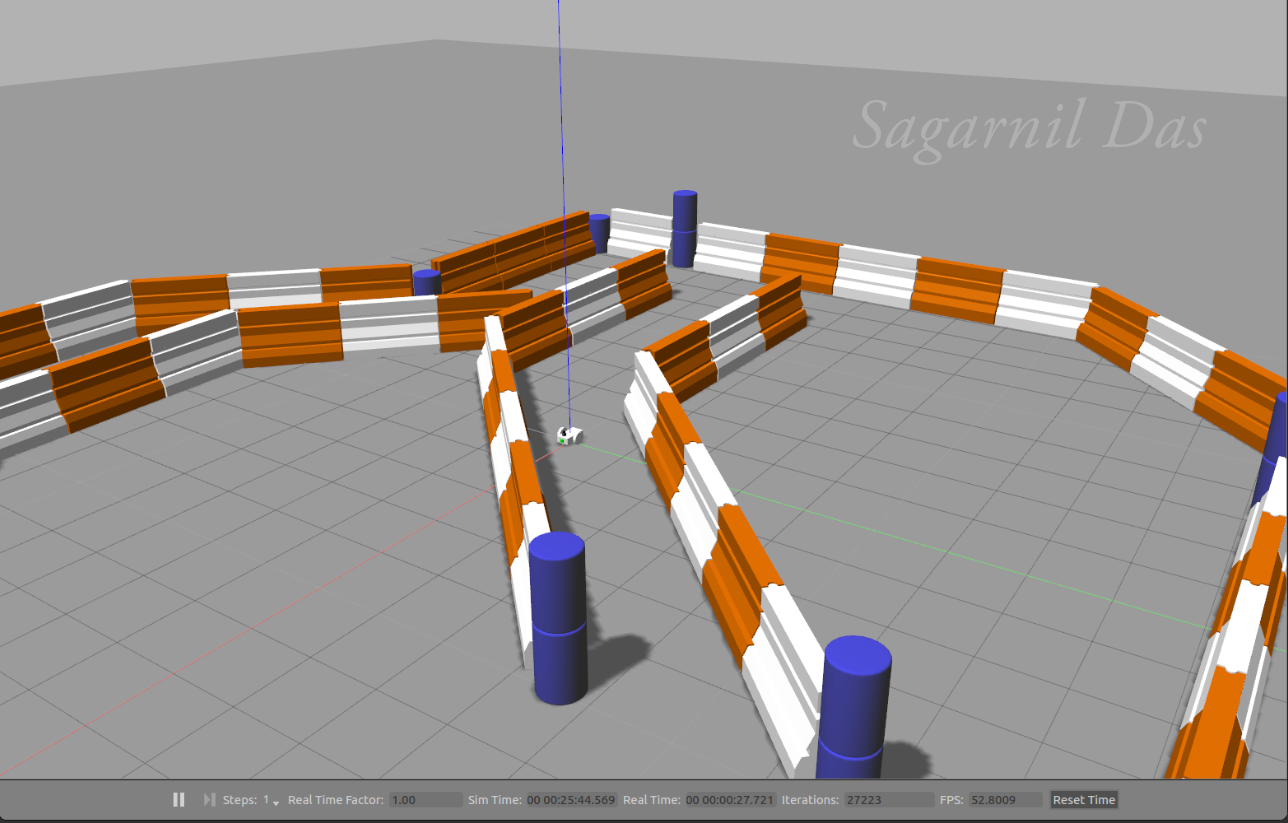}
      \caption{Robot in Gazebo environment}
      \label{fig:Robot in Gazebo environment}
\end{figure}

Initially, for both the bots, particles are very dispersed indicating great uncertainty in the robot position. At this point, sensors have not yet provided any information regarding the location. Fig 5 shows the great uncertainty of the robot's position represented by the particles denoted by red arrows.

\begin{figure}[thpb]
      \centering
      \includegraphics[width=\linewidth]{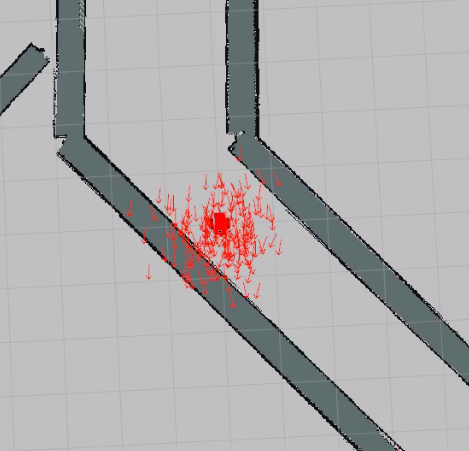}
      \caption{High uncertainty in robot position}
      \label{fig:High uncertainty in robot position}
\end{figure}

After the simulation is started, the localization process starts from taking the sensor measurements and gradually improve. After the algorithm converges, the particles effectively depicts the pose of the robot in the map, thus making the robot successfully navigate through the maze and reach the goal state.

\subsection{Achievements}
For this project, two robots were deployed in the simulation environment. The benchmark model or \textit{UdacityBot} and the custom made model or \textit{SagarBot}. 

\subsubsection{UdacityBot}

The UdacityBot initially started towards the north of the map as from its local costmap, it calculated the path to the goal from its starting position to be shorter than any other paths. But it soon discovered the presence of an obstacle and the impossibility of reaching the goal through that path. Then it turned around and reached the goal through the 2nd most shortest path. Fig. 6 depicts UdacityBot at the goal position.

\begin{figure}[thpb]
      \centering
      \includegraphics[width=\linewidth]{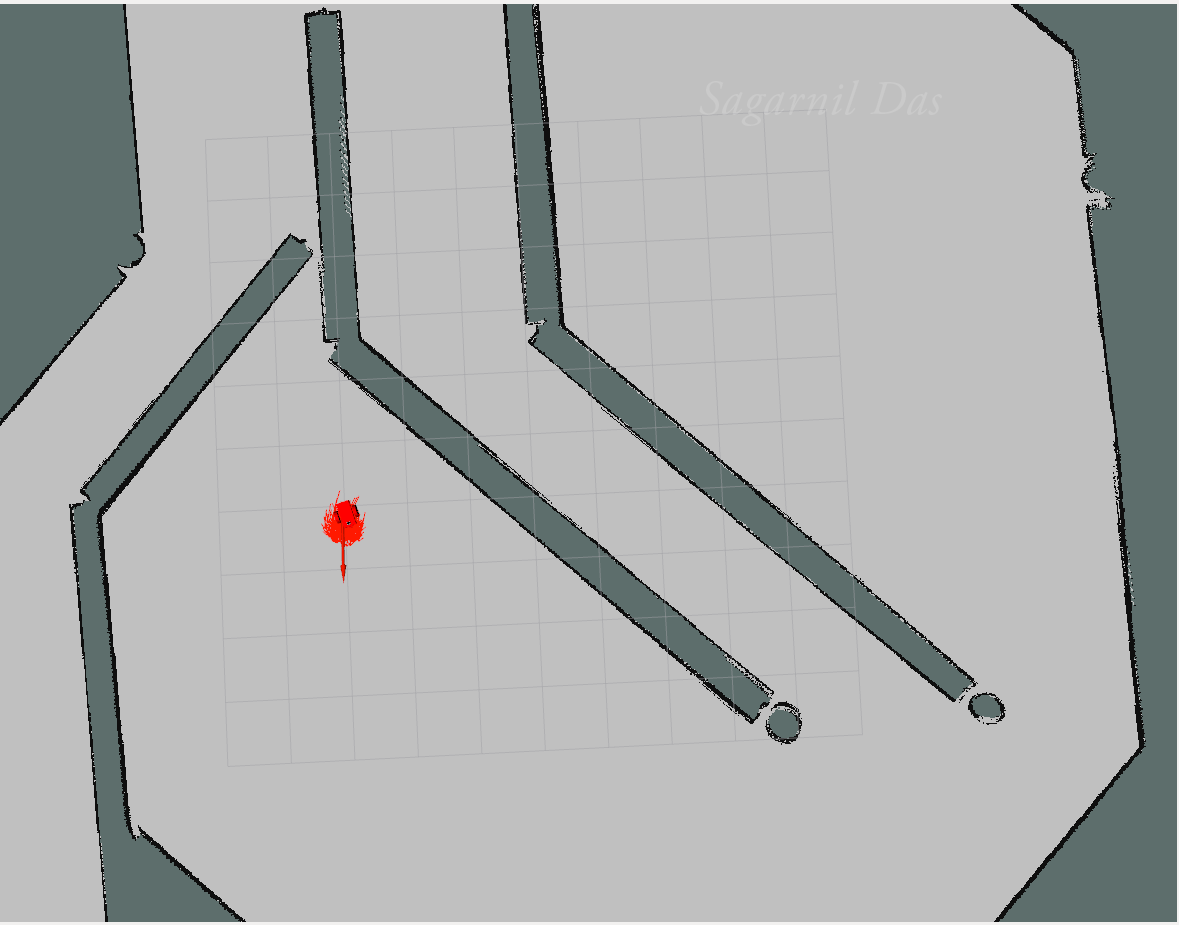}
      \caption{UdacityBot at the goal position}
      \label{fig:UdacityBot at the goal position}
\end{figure}

\subsubsection{SagarBot}

The SagarBot also exhibited similar behavior and started towards north of the map, before discovering the impossibility of that route and then it reached the goal position through more or less the same route UdacityBot has taken. Fig. 7 depicts SagarBot at the goal position.

\begin{figure}[thpb]
      \centering
      \includegraphics[width=\linewidth]{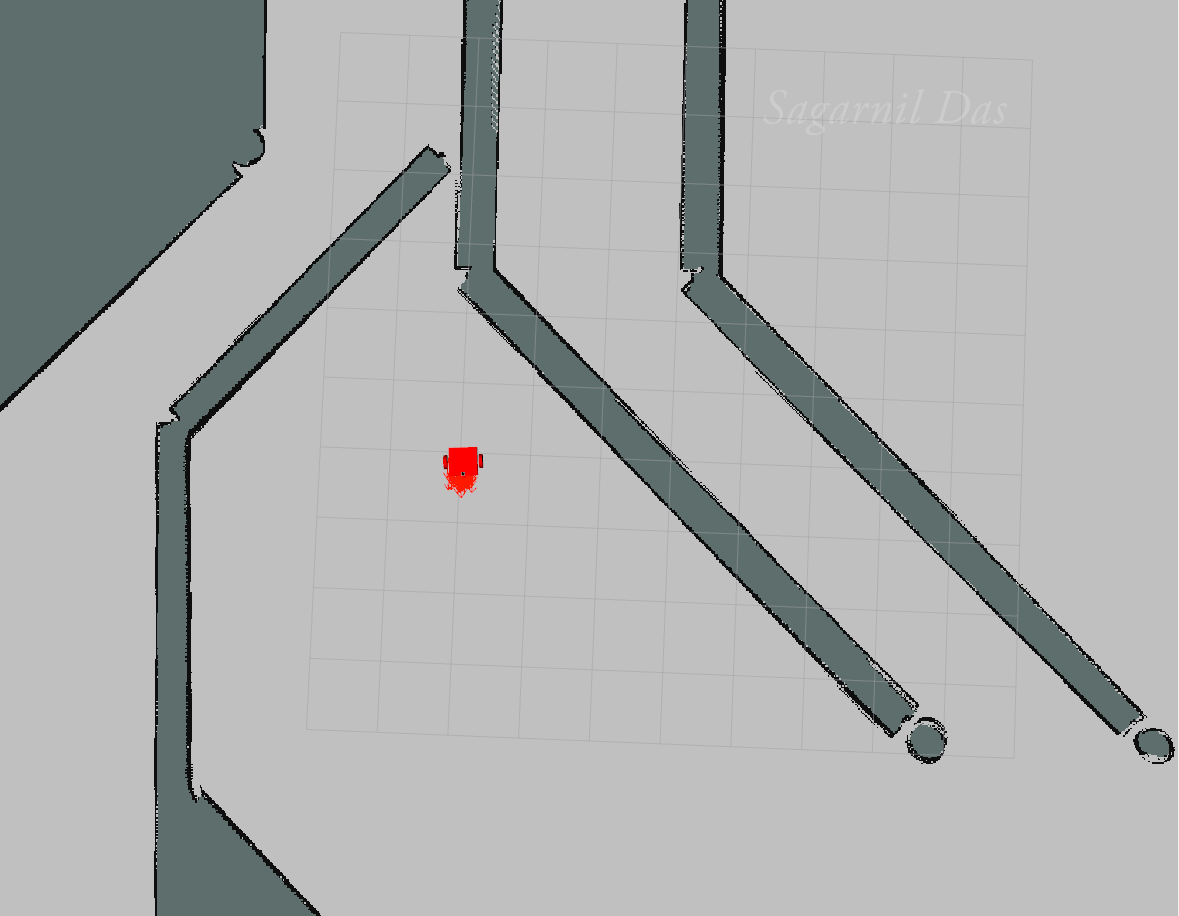}
      \caption{SagarBot at the goal position}
      \label{fig:SagarBot at the goal position}
\end{figure}

\subsection{Benchmark Model}
\subsubsection{Model design}
The robot's design considerations included the size of the robot and the layout of the sensors. They are discussed below.
\paragraph{Maps}

The Clearpath\cite{jackalrace} jackalrace.yaml and jackalrace.pgm were used to create the maps.

\paragraph{Meshes}

The laser scanner which was used in the robot for detecting obstacles is the Hokuyo scanner\cite{Hokuyo_scanner}. The mesh \textit{hokuyo.dae} was used to render it.

\paragraph{Launch files}

Three launch files were used. They are as follows:

1. \textbf{\texttt{robot\_description.launch}}: This launch file defines the \textit{\texttt{joint\_state\_publisher}} which sends fake joint values, \textit{\texttt{robot\_state\_publisher}} which sends robot states to tf and \textit{\texttt{robot\_description}} which defines and sends the URDF to the parameter server.

2. \textbf{amcl.launch}: The amcl package relies entirely on the robot’s odometry and the laser scan data. This file launches the AMCL localization server, the map server, the odometry frame, the \texttt{move\_base} server and the trajectory planner server.

3. \textbf{\texttt{udacity\_world.launch}}: This is the primary launch file which contains the \texttt{robot\_description.launch} file, the gazebo world and the AMCL localization server. It also spawns the robot and launches RViz. Fig. 8 depicts the connection graph between the discussed nodes. Fig. 9 depicts the UdacityBot at the goal location and the AMCL particles as seen in RViz.

\begin{figure}[thpb]
      \centering
      \includegraphics[width=\linewidth]{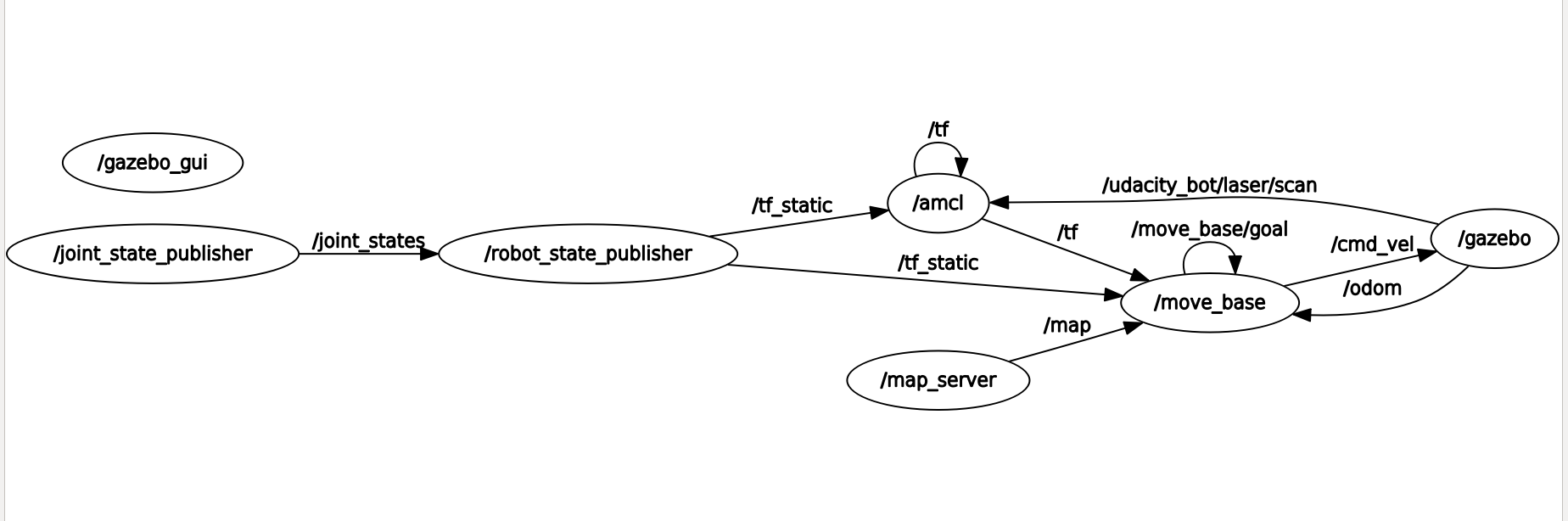}
      \caption{Node relations}
      \label{fig:Node relations}
\end{figure}

\begin{figure}[thpb]
      \centering
      \includegraphics[width=\linewidth]{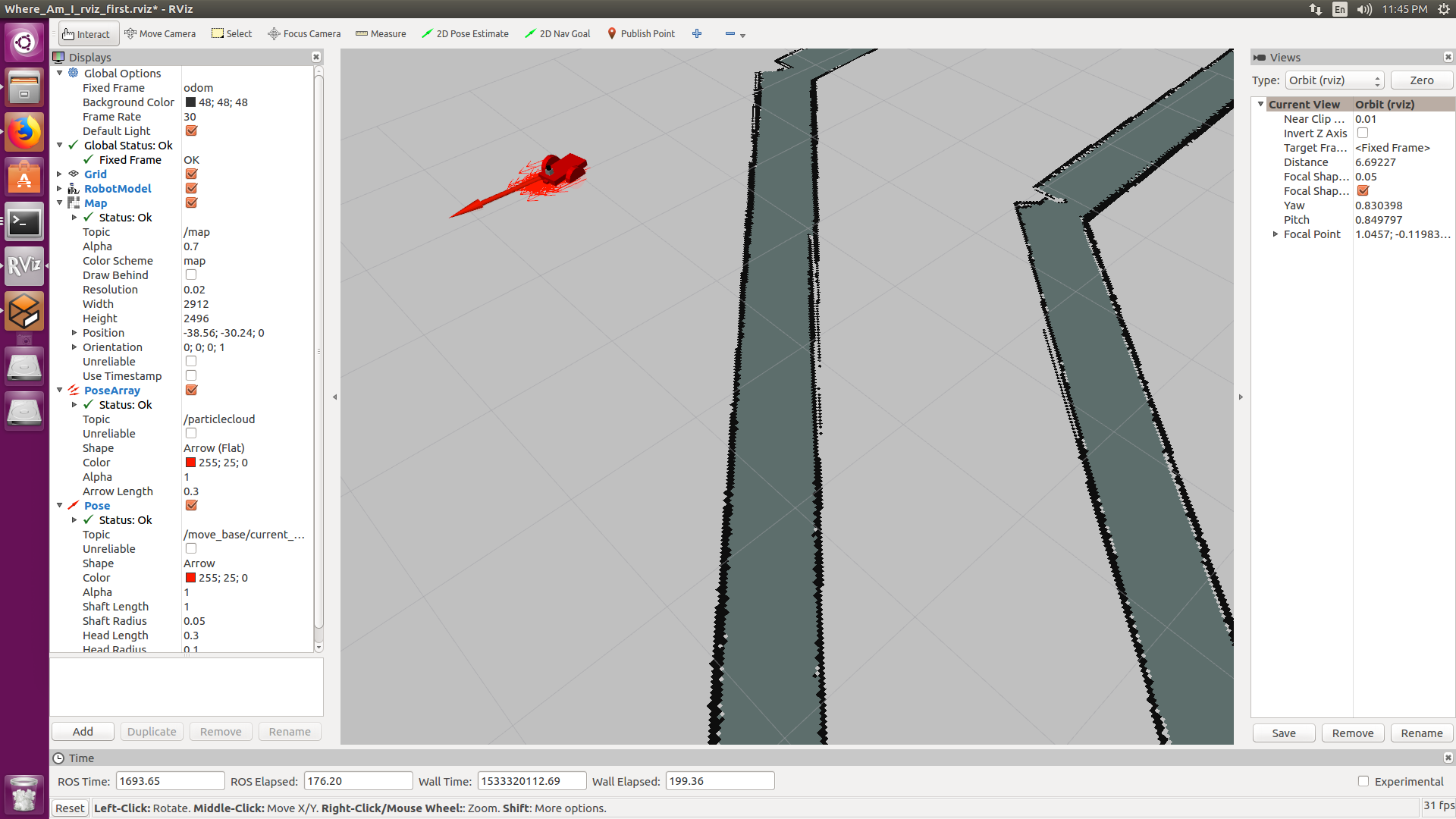}
      \caption{RViz interface with UdacityBot at goal position}
      \label{fig:RViz interface with UdacityBot at goal position}
\end{figure}

\paragraph{Worlds}

This project was done in two worlds.

1. \textbf{Udacity world}: This is the original blank world where the robots were created and prototyped. This defines the ground plane, the light source and the world camera.

2. \textbf{\texttt{jackal\_race world}}: This world defines the maze.

\paragraph{URDF}

The URDF files defines the shape and size of the robot. Two files were used as URDFs:

1. \textbf{\texttt{udacity\_bot.xacro}}: Provides the shape and size of the robot in macro format. For the UdacityBot, a fixed base is used. A single link, with the name defined as "chassis", encompassed the base as well as the caster wheels. Each link has specific elements, such as the inertial or the collision elements. The chassis is a cuboidal (or box), whereas the casters are spherical as denoted by their "geometry" tags. Each link (or joint) has an origin (or pose) defined as well. Every element of that link or joint will have its own origin, which will be relative to the link's frame of reference.

For this base, as the casters are included as part of the link (for stability purposes), there is no need for any additional links to define the casters, and therefore no joints to connect them. The casters do, however, have friction coefficients defined for them, and are set to 0, to allow for free motion while moving. 

Two wheels were attached to the robot. Each wheel is represented as a link and is connected to the base link (the chassis) with a joint. For each wheel, a "collision", "inertial" and "visual" elements are present. The joint type is set to "continuous" and is similar to a revolute joint but has no limits on its rotation. It can rotate continuously about an axis. The joint will have it's own axis of rotation, some specific joint dynamics that correspond to the physical properties of the joint like "friction", and certain limits to enforce the maximum "effort" and "velocity" for that joint. The limits are useful constraints in regards to a physical robot and can help create a more robust robot model in simulation as well.

For the UdacityBot, two sensors were used. A camera and a Laser range-finder (hokuyo sensor).

2. \textbf{\texttt{udacity\_bot.gazebo}}: This file was included as the URDF file is unable to make the robot take pictures with the camera or detect obstacle with the Laser range-finder. This file contains 3 plugins, one each for the camera sensor, the hokuyo sensor and the wheel joints. It also implements a differential drive controller.

Fig. 10 depicts the navigation goal messages.

\begin{figure}[thpb]
      \centering
      \includegraphics[width=\linewidth]{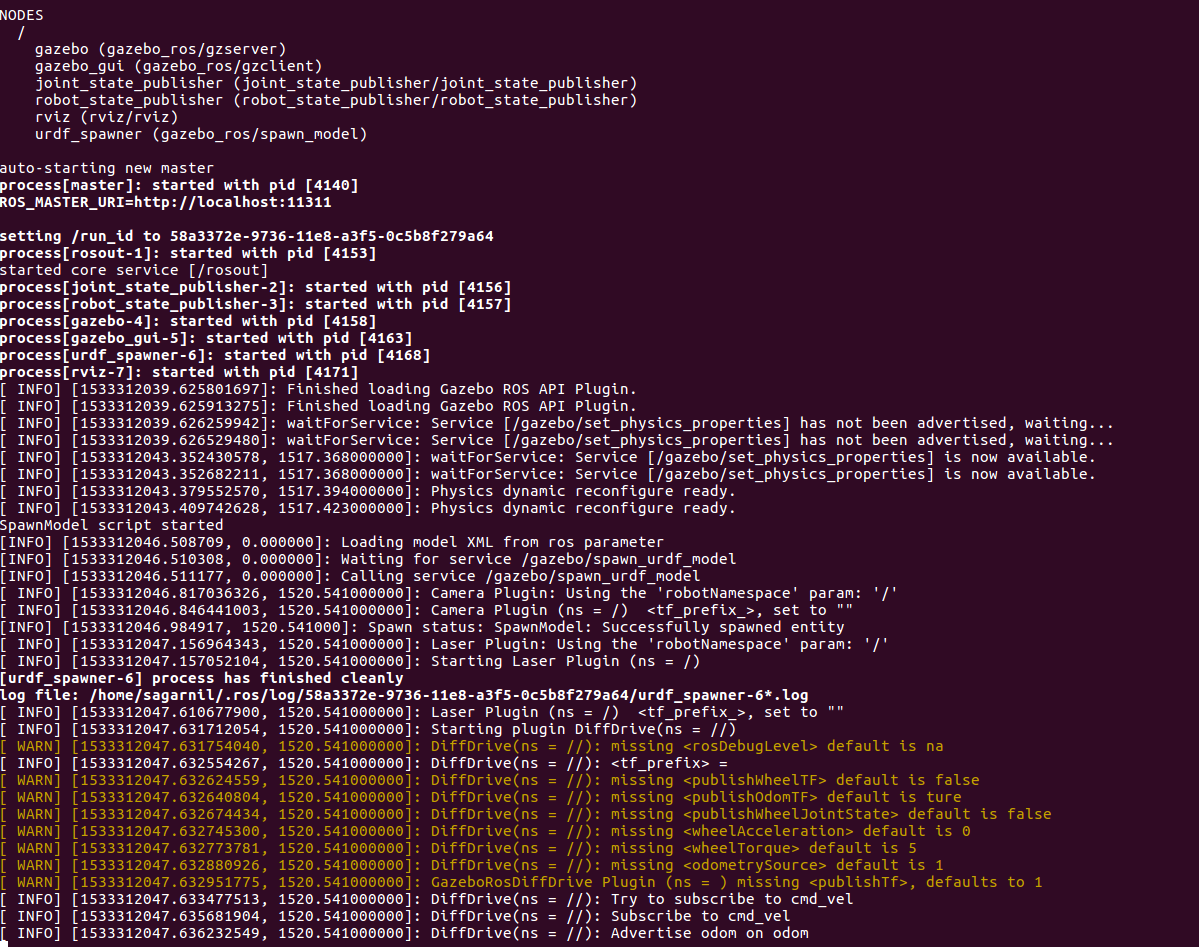}
      \caption{Navigation goal messages}
      \label{fig:Navigation goal messages}
\end{figure}

\subsubsection{Packages Used}
A ros package called \texttt{udacity\_bot} was designed for this project. The structure of this package is shown below.

\begin{itemize}
\item config
\item images
\item launch 
\item maps
\item meshes
\item src
\item urdf
\item worlds
\end{itemize}

This package, along with the AMCL and the navigation stack packages were crucial for the success of the mobile robot in performing a localization task. Table 1 describes UdacityBot setup instructions.

\begin{table}[h]
\caption{Udacity Bot setup instructions}
\begin{tabular}{@{}|c|l|c|l|c|l|@{}}
\toprule
\multicolumn{6}{|c|}{\textbf{Udacity Bot Body}}                                                                                                                                                                                                                                                                                               \\ \midrule
\multicolumn{2}{|c|}{\textbf{Part}}          & \multicolumn{2}{c|}{\textbf{Geometry}}                                                                                          & \multicolumn{2}{c|}{\textbf{Size}}                                                                                                                           \\ \midrule
\multicolumn{2}{|c|}{Chassis}          & \multicolumn{2}{c|}{Cube}                                                                                                       & \multicolumn{2}{c|}{0.4 x 0.2 x 0.1}                                                                                                                         \\ \midrule
\multicolumn{2}{|l|}{Back and Front Casters} & \multicolumn{2}{c|}{Sphere}                                                                                                     & \multicolumn{2}{c|}{0.0499 (radius)}                                                                                                                         \\ \midrule
\multicolumn{2}{|c|}{Left and Right wheels}  & \multicolumn{2}{c|}{Cylinders}                                                                                                  & \multicolumn{2}{l|}{0.1 (radius), 0.05 (length)}                                                                                                             \\ \midrule
\multicolumn{2}{|c|}{Camera Sensor}          & \multicolumn{2}{c|}{\begin{tabular}[c]{@{}c@{}}Link Origin\\ Shape-size\\ Joint Origin\\ Parent Link\\ Child Link\end{tabular}} & \multicolumn{2}{c|}{\begin{tabular}[c]{@{}c@{}}{[}0, 0, 0, 0, 0, 0{]}\\ Box - 0.05 x 0.05 x 0.05\\ {[}0.2, 0, 0, 0, 0, 0{]}\\ chassis\\ camera\end{tabular}} \\ \midrule
\multicolumn{2}{|c|}{Hokuyo Sensor}          & \multicolumn{2}{l|}{\begin{tabular}[c]{@{}l@{}}Link Origin\\ Shape-Size\\ Joint Origin\\ Parent Link\\ Child Link\end{tabular}} & \multicolumn{2}{c|}{\begin{tabular}[c]{@{}c@{}}{[}0, 0, 0, 0, 0, 0{]}\\ Box - 0.1 x 0.1 x 0.1\\ {[}0.15, 0, 0.1, 0, 0, 0{]}\\ chassis\\ hokuyo\end{tabular}} \\ \bottomrule
\end{tabular}
\end{table}

\subsubsection{Parameters}
Exploring, adding and tuning parameters for the AMCL and move\_base packages were some of the main goals of this project. For effective parameter tuning, ROS basic navigation tuning guide\cite{ros_navigation_tuning} and Kaiyu Zheng's navigation tuning guide\cite{Kaiyu_Zheng_navigation_tuning} were heavily used. The parameters were iteratively tuned to see what works best for the UdacityBot. The AMCL parameters were tuned as follows:

The min and max particles parameters were set to 25 and 200 in order to prevent over-usage of CPU. Increasing the max\_particles did not improve the robot's initial ability to find itself with certainty.

The transform\_tolerance was one of the main parameters to tune. The tf package, helps keep track of multiple coordinate frames, such as the transforms from these maps, along with any transforms corresponding to the robot and its sensors. Both the amcl and move\_base packages or nodes require that this information be up-to-date and that it has as little a delay as possible between these transforms. The maximum amount of delay or latency allowed between transforms is defined by the transform\_tolerance parameter. It was finally set to 1.25 for the cost-maps and 0.2 for the amcl package.

The laser model parameters like laser\_max\_beams, laser\_z\_hit and laser\_z\_rand, were kept as default as the obstacles were clearly detected in the local cost-maps with them as is. Fig. 11 depicts this observation.

\begin{figure}[thpb]
      \centering
      \includegraphics[width=\linewidth]{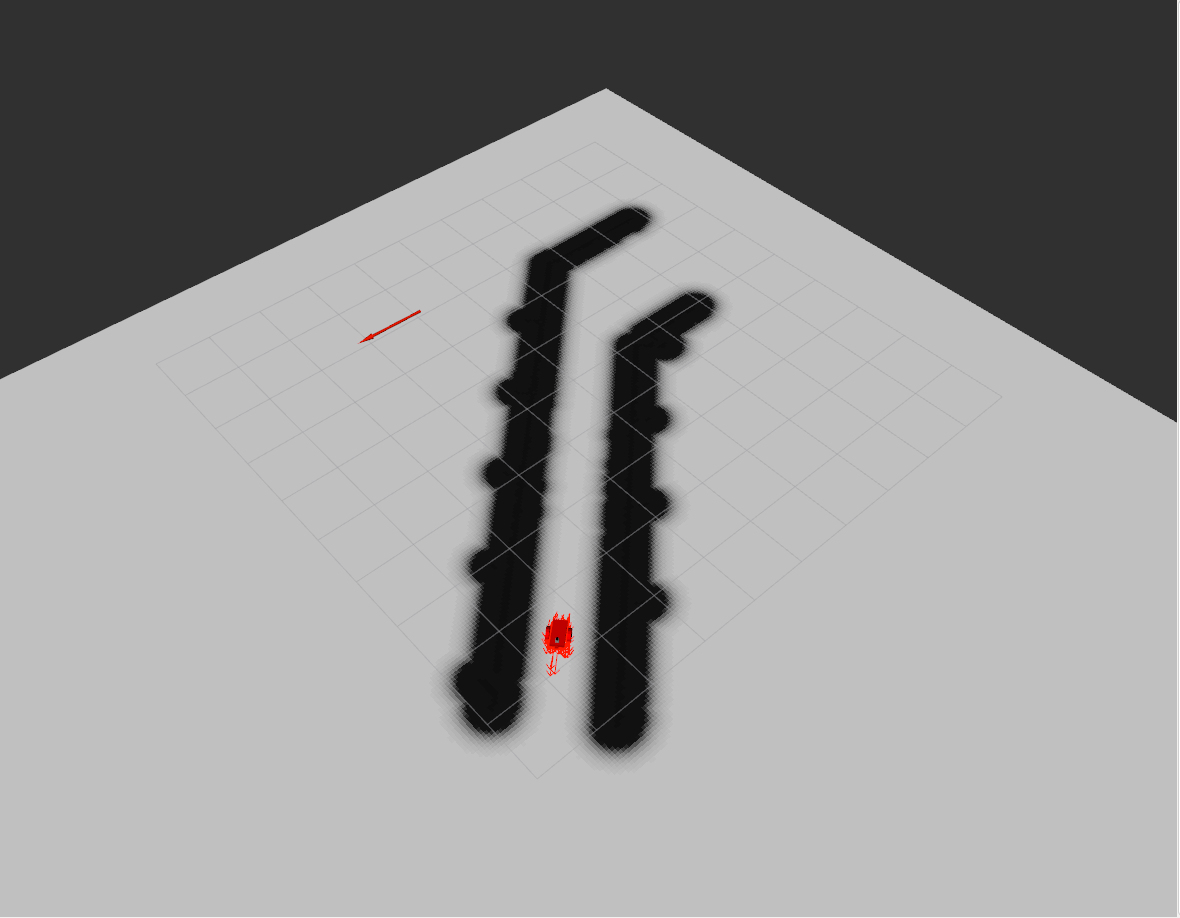}
      \caption{Local cost-map with AMCL particles converged}
      \label{Local cost-map}
\end{figure}

The odom\_model\_type was kept as \textit{diff-corrected} as this mobile robot followed a differential drive. There are additional parameters that are specific to this type - the odom\_alphas (1 through 4). These parameters define how much noise is expected from the robot's movements/motions as it navigates inside the map. They were kept at 0.005, 0.005, 0.010 and 0.005 respectively obtained from trial-error method. Table 2 depicts the AMCL parameters used.

\begin{table}[h]
\caption{AMCL parameters}
\begin{tabular}{@{}|c|c|@{}}
\toprule
\multicolumn{2}{|c|}{AMCL Parameters}                \\ \midrule
Parameter                    & Value                 \\ \midrule
odom\_frame\_id              & odom                  \\ \midrule
odom\_model\_type            & diff-corrected        \\ \midrule
transform\_tolerance         & 0.2                   \\ \midrule
min\_particles               & 25                    \\ \midrule
max\_particles               & 200                   \\ \midrule
initial\_pose\_x             & 0.0                   \\ \midrule
initial\_pose\_y             & 0.0                   \\ \midrule
initial\_pose\_a             & 0.0                   \\ \midrule
laser\_z\_hit                & 0.95                  \\ \midrule
laser\_z\_short              & 0.1                   \\ \midrule
laser\_z\_max                & 0.05                  \\ \midrule
laser\_z\_rand               & 0.5                   \\ \midrule
laser\_sigma\_hit            & 0.2                   \\ \midrule
laser\_lambda\_short         & 0.1                   \\ \midrule
laser\_model\_type           & likelihood\_field     \\ \midrule
laser\_likelihood\_max\_dist & 2.0                   \\ \midrule
odom\_alpha1                 & 0.005                 \\ \midrule
odom\_alpha2                 & 0.005                 \\ \midrule
odom\_alpha3                 & 0.010                 \\ \midrule
odom\_alpha4                 & 0.005                 \\ \midrule
\end{tabular}
\end{table}

The move\_base parameters were tuned as follows:

The obstacle\_range parameter was modified to have a greater value of 1.5. This parameter depicts the default maximum distance from the robot (in meters) at which an obstacle will be added to the cost-map.

The raytrace\_range parameter was modified to a higher value of 4.0. This parameter is used to clear and update the free space in the cost-map as the robot moves. 

Two parameters in the global and local cost-maps were also changed. They are:

1. update\_frequency: This value was set to 10.0. This is the frequency in Hz for the map to be updated.

2. publish\_frequency: This value was also set to 10.0. This is the frequency at which the map will be published on the display. 

The yaw\_goal\_tolerance was updated to a value of 0.1. This parameter depicts the tolerance in radians for the controller in yaw/rotation when achieving its goal. The xy\_goal\_tolerance was updated to a value of 0.2. This is the tolerance in meters for the controller in the x-y distance when achieving the goal. Both these parameters were doubled to allow for additional flexibility in trajectory planning. The transform\_tolerance was set to 1.25. Table 3 depicts the move\_base parameters used.

\begin{table}[h]
\caption{move\_base parameters}
\begin{tabular}{@{}|c|c|@{}}
\toprule
\multicolumn{2}{|c|}{move\_base Parameters} \\ \midrule
Parameter                    & Value        \\ \midrule
yaw\_goal\_tolerance         & 0.1          \\ \midrule
xy\_goal\_tolerance          & 0.2          \\ \midrule
obstacle\_range              & 1.5          \\ \midrule
raytrace\_range              & 4.0          \\ \midrule
inflation\_radius            & 0.65         \\ \midrule
robot\_radius                & 0.3          \\ \midrule
update\_frequency            & 10.0         \\ \midrule
publish\_frequency           & 10.0         \\ \bottomrule
\end{tabular}
\end{table}

\subsection{Personal Model - SagarBot}
\subsubsection{Model design}

The SagarBot has a similar structure as UdacityBot, but it has a square base and is bigger in size. The laser sensor was also moved to the front of the robot.

\paragraph{Maps}

The Clearpath \textit{jackal\_race.yaml} and \textit{jackal\_race.pgm} packages were used to create the map, similar to the UdacityBot.\cite{jackalrace}

\paragraph{Meshes}

A hokuyo scanner was simulated as the laser scanner. The hokuyo.dae mesh was used to render it.

\paragraph{Launch}

Three launch files were used. They are:

1. \textbf{\texttt{robot\_description\_sagar.launch}}: This launch file defines the \textit{\texttt{joint\_state\_publisher}} which sends fake joint values, \textit{\texttt{robot\_state\_publisher}} which sends robot states to tf and \textit{\texttt{robot\_description}} which defines and sends the URDF to the parameter server.

2. \textbf{\texttt{amcl\_sagar.launch}}: The amcl package relies entirely on the robot’s odometry and the laser scan data. This file launches the AMCL localization server, the map server, the odometry frame, the \texttt{move\_base} server and the trajectory planner server.

3. \textbf{\texttt{udacity\_world\_sagar.launch}}: This is the primary launch file which contains the \texttt{robot\_description\_sagar.launch} file, the gazebo world and the AMCL localization server. It also spawns the robot and launches RViz. Fig. 8 depicts the connection graph between the discussed nodes. Fig. 9 depicts the UdacityBot at the goal location and the AMCL particles as seen in RViz.

\paragraph{Worlds}: The SagarBot uses the same worlds as the UdacityBot.

\paragraph{URDF}: The URDF files defines the shape and size of the robot. Two files were used to define the basic robot description and the gazebo view.

1. \textbf{sagar\_bot.gazebo}: This file defines the differential drive controller, the camera and the camera controller, the controller for Gazebo and Hokuyo laser scanner.

2. \textbf{sagar\_bot.xacro}: Defines the robot's shape in macro format. sagar\_bot follows a similar structure as udacity\_bot except it's square in shape and bigger. Table 4 defines the parameters used in sagar\_bot.xacro.

\begin{table}[h]
\caption{SagarBot setup instructions}
\begin{tabular}{@{}|c|l|c|l|c|l|@{}}
\toprule
\multicolumn{6}{|c|}{\textbf{SagarBot Body}}                                                                                                                                                                                                                                                                                               \\ \midrule
\multicolumn{2}{|c|}{\textbf{Part}}          & \multicolumn{2}{c|}{\textbf{Geometry}}                                                                                          & \multicolumn{2}{c|}{\textbf{Size}}                                                                                                                           \\ \midrule
\multicolumn{2}{|c|}{Chassis}          & \multicolumn{2}{c|}{Cube}                                                                                                       & \multicolumn{2}{c|}{0.4 x 0.4 x 0.1}                                                                                                                         \\ \midrule
\multicolumn{2}{|l|}{Back and Front Casters} & \multicolumn{2}{c|}{Sphere}                                                                                                     & \multicolumn{2}{c|}{0.05 (radius)}                                                                                                                         \\ \midrule
\multicolumn{2}{|c|}{Left and Right wheels}  & \multicolumn{2}{c|}{Cylinders}                                                                                                  & \multicolumn{2}{l|}{0.1 (radius), 0.05 (length)}                                                                                                             \\ \midrule
\multicolumn{2}{|c|}{Camera Sensor}          & \multicolumn{2}{c|}{\begin{tabular}[c]{@{}c@{}}Link Origin\\ Shape-size\\ Joint Origin\\ Parent Link\\ Child Link\end{tabular}} & \multicolumn{2}{c|}{\begin{tabular}[c]{@{}c@{}}{[}0, 0, 0, 0, 0, 0{]}\\ Box - 0.05 x 0.05 x 0.05\\ {[}0.2, 0, 0, 0, 0, 0{]}\\ chassis\\ camera\end{tabular}} \\ \midrule
\multicolumn{2}{|c|}{Hokuyo Sensor}          & \multicolumn{2}{l|}{\begin{tabular}[c]{@{}l@{}}Link Origin\\ Shape-Size\\ Joint Origin\\ Parent Link\\ Child Link\end{tabular}} & \multicolumn{2}{c|}{\begin{tabular}[c]{@{}c@{}}{[}0, 0, 0, 0, 0, 0{]}\\ Box - 0.1 x 0.1 x 0.1\\ {[}0.15, 0, 0.1, 0, 0, 0{]}\\ chassis\\ hokuyo\end{tabular}} \\ \bottomrule
\end{tabular}
\end{table}

\subsubsection{Packages Used}

Same as UdacityBot.

\subsubsection{Parameters}

The AMCL parameters were kept the same as UdacityBot, but significant changes were made in the move\_base parameters. The move\_base parameters were tuned as follows:

The obstacle\_range parameter was modified to have a greater value of 5.0. This parameter depicts the default maximum distance from the robot (in meters) at which an obstacle will be added to the cost-map. This was done in a trial-and-error method. The basic intuition behind increasing this parameter was as the SagarBot was larger in size, it moved slower. So additional time was wasted if it was unable to see an obstacle in moderate range and followed the path towards it only to find the region bounded by an obstacle. Hence, in order to increase its sight with respect to the cost-map, a higher value of this parameter was used.

The raytrace\_range parameter was modified to a higher value of 8.0. This parameter is used to clear and update the free space in the cost-map as the robot moves. 

Two parameters in the global and local cost-maps were also changed. They are:

1. update\_frequency: This value was set to 10.0. This is the frequency in Hz for the map to be updated.

2. publish\_frequency: This value was also set to 10.0. This is the frequency at which the map will be published on the display. 

The yaw\_goal\_tolerance was updated to a value of 0.1. This parameter depicts the tolerance in radians for the controller in yaw/rotation when achieving its goal. The xy\_goal\_tolerance was updated to a value of 0.2. This is the tolerance in meters for the controller in the x-y distance when achieving the goal. Both these parameters were doubled to allow for additional flexibility in trajectory planning. The transform\_tolerance was set to 1.25. Table 5 depicts the move\_base parameters used.

\begin{table}[h]
\caption{move\_base parameters for SagarBot}
\begin{tabular}{@{}|c|c|@{}}
\toprule
\multicolumn{2}{|c|}{move\_base Parameters} \\ \midrule
Parameter                    & Value        \\ \midrule
yaw\_goal\_tolerance         & 0.1          \\ \midrule
xy\_goal\_tolerance          & 0.2          \\ \midrule
obstacle\_range              & 5.0         \\ \midrule
raytrace\_range              & 8.0          \\ \midrule
inflation\_radius            & 0.55         \\ \midrule
robot\_radius                & 0.4          \\ \midrule
update\_frequency            & 10.0         \\ \midrule
publish\_frequency           & 10.0         \\ \bottomrule
\end{tabular}
\end{table}

\section{Results}

\subsection{Localization Results}
\subsubsection{Benchmark - UdacityBot}

The time taken for the particle filters to converge was around 5-6 seconds. The UdacityBot reaches the goal within approximately two minutes. So the localization results are pretty decent considering the time taken for the localization and reaching the goal. However, it does not follow a smooth path for reaching the goal. Initially, the robot heads towards the north as it was unable to add the obstacle over there in its local cost-map and hence it followed the shortest path to the goal. But soon, it discovered the presence of the obstacle and it changed its strategy to the next shortest route to reach the goal, i.e. head south-east, turn around where the obstacle ends and reach the goal.

\subsubsection{Student - SagarBot}

The time taken for the particle filters to converge was around 30-40 seconds. The SagarBot reaches the goal within approximately 15-20 minutes. So here, a deterioration of the results is observed. This can be attributed to the heavier mass of the SagarBot, even though no wheel slippage was kept for both the robots in their respective URDFs.

\subsection{Technical Comparison} 
SagarBot was significantly heavier than the UdacityBot. SagarBot has a square shape and also the Laser finder was moved to the front.

\section{Discussion}

UdacityBot performed significantly better than SagarBot. This might be attributed to the fact that SagarBot is considerably heavier than UdacityBot. This significantly changed SagarBot's speed and hence the time it takes to reach the goal also increased. Sagarbot, which occupies more space than UdacityBot also had a problem navigating with a higher inflation\_radius as it thought the obstacles to be thicker than they really are and hence deducing that it has not enough space to navigate. As a result, sometimes SagarBot froze in one place trying to figure out an alternate route which took a long time. So, when the inflation\_radius was decreased, the performance of SagarBot improved considerably. 

The obstacle\_range was also another important parameter to tune. SagarBot, being slower in speed naturally wastes a lot of time if it goes to the wrong direction only to find an obstacle there. So in order to be able to successfully add an obstacle to its cost-map from a significantly larger distance, this parameter was increased which resulted in significant improvement of SagarBot's performance.

\subsection{Topics}
\begin{itemize}
\item UdacityBot performed better.
\item One might infer the better performance of UdacityBot due to its smaller mass.
\item In the 'Kidnapped-robot' problem, one has to successfully account for the scenario that at any point in time, the robot might be kidnapped and placed in a totally different position of the world.
\item In any scenario, where the world map is known, but the position of the robot is not, localization can be successfully used. 
\item MCL/AMCL can work well in any industry domain where the robot's path is guided by clear obstacles and where the robot is supposed to reach the goal state from anywhere in the map. The ground also needs to be flat and obstacle free, particularly in the cases where the laser range-finder is at a higher position and cannot detect objects on the ground.
\end {itemize}

\section{Conclusion / Future work}

Both robots satisfied the two conditions, i.e. Both were successfully localized with the help of AMCL algorithm and both reached the goal state within a decent time-limit. UdacityBot performed better than SagarBot which can be attributed to the heavier mass of SagarBot. This time difference was reduced significantly by increasing the obstacle\_range and inflation\_radius parameters. The main issue with both the robots was erroneous navigation as both of the robots took a sub-optimal route due to the lack of knowledge of an obstacle beforehand. However, as the primary focus of this project was localization, this problem was ignored due to time constraint.

Even though both the robots reached the goal eventually, as they failed to take the optimal route to the goal location, this implies that neither of these robots can be deployed in commercial products.

Placement of the laser scanner can also play a vital role in the robot's navigation skill. Placing the scanner too high from the ground could result in the robot missing obstacles in the ground and then get stuck on it causing a total sensor malfunction. On the other hand, placing the scanner too low may prevent the robot from perceiving better as it gets more viewing range in this situation. So this would be another important modification to improve the robots.

Future work would involve making both the robots commercially viable by working on making/tuning a better navigation planner.

Also, the present model deployed only has one robot in a world at a time. In future, this project could be expanded to include multiple robots in the same world, each with the same goal or a different one.

For a detailed version of this project, this github reopsitory can be referred to: https://github.com/sagarnildass/Robotics\_Nanodegree/tree/master/Term2/Project\_2\_RoboND\_Where\_am\_I/udacity\_bot.\cite{github}

\subsection{Hardware Deployment}
\begin{enumerate}
\item This present model was done in a simulation environment in a local computer hosting Ubuntu 16.04 on a core-i7 machine and NVIDIA GTX 1080Ti GPU. In order to deploy it in a hardware system like NVIDIA Jetson TX2, the TX2 prototype board camera could be connected into the model with suitable drivers. Laser range-finder hardwares would also need to be integrated in order the hardware version to successfully operate and detect obstacles. It would also need GIO connections for driving the wheels and be implemented on a suitable platform modeling the robot which was simulated.
\item The JETSON TX2 has sufficient has adequate processing power both in CPU and GPU memory to successfully host this model.
\end{enumerate}

\bibliography{bib}

\begin{thebibliography}{1}

\bibitem{jackalrace}
C.~Robotics, ``Navigating with jackal,''

\bibitem{particle_filters}
S.~Thrun, {\em Particle filters in Robotics}.
\newblock Carnegie Mellon University, 2002.

\bibitem{kalman_filters}
Q.~Li~et al, ``Kalman filters and its applications,'' 2015.

\bibitem{navigation_stack}
M.~Quigley~et al, ``Ros: an open-source robot operating system,'' 2009.

\bibitem{Hokuyo_scanner}
https://autonomoustuff.com/product-category/lidar/hokuyo-laser scanners/,
  ``Hokuyo scanner home page,'' 2018.

\bibitem{ros_navigation_tuning}
http://wiki.ros.org/navigation/Tutorials/Navigation
  basic navigation tuning guide,'' 2018.

\bibitem{Kaiyu_Zheng_navigation_tuning}
K.~Zheng, ``Kaiyu zheng basic navigation tuning guide,'' 2016.

\bibitem{github}
S.~Das, ``Robotic localization in a mapped environment,'' 2018.

\end{thebibliography}
\bibliographystyle{ieeetr}

\end{document}